\documentclass[cameraready]{Interspeech}
\title{Said Aloud, Read Different: Cross-Modal Instability in Multimodal Models}
\author[]{Basel}{Mousi}
\author[]{Fahim}{Dalvi}
\author[]{Shammur}{Chowdhury}
\author[]{Firoj}{Alam}
\author[]{Nadir}{Durrani}

\address{
     Qatar Computing Research Institute, HBKU, Qatar 
}

\email{\{bmousi,faimaduddin,shchowdhury,fialam,ndurrani\}@hbku.edu.qa}


\usepackage{comment}
\usepackage{amssymb}
\usepackage[utf8]{inputenc} 
\usepackage[T1]{fontenc}
\usepackage{fontawesome5}
\usepackage{graphicx}
\usepackage{xcolor}
\usepackage{soul}
\usepackage[normalem]{ulem}

\usepackage{tabularx}
\usepackage{booktabs}
\usepackage{array}
\usepackage{textcomp}

\definecolor{tri}{rgb}{.25,.88,.82}
\definecolor{lilac}{rgb}{0.85,0.64,0.85}
\definecolor{lightblue}{rgb}{.50,.95,1}

\begin{document}

\maketitle

\begin{abstract}


Multimodal foundation models are increasingly used in speech-first assistants that must interpret spoken queries and produce visually grounded decisions. Yet it remains unclear whether semantically equivalent queries yield consistent judgments across modality (text \emph{vs.} speech) and language (English vs. Arabic). We introduce a \textit{speech-augmented visually grounded contrastive triplet benchmark} spanning \textit{10{,}150} culturally grounded images from \textit{18 MENA countries}, where each image is paired with one supported statement and two plausible but unsupported alternatives. We define contrastive instability as the conditional rate at which a model fails to resolve all statements within a triplet, isolating fragmented reasoning from complete failure. Evaluating recent multimodal models under text and speech in English and Arabic, we find that modality and language shifts introduce substantial triplet-level inconsistencies that are not fully captured by aggregate accuracy, with speech amplifying partial failures. We  make the benchmark publicly available to the community.\footnote{\url{https://huggingface.co/datasets/QCRI/M2CQA-S}}

\end{abstract}

\section{Introduction}
\label{sec:intro}

Multimodal foundation models are increasingly accessed through speech interfaces \cite{radford2022robust,zhang-etal-2023-speechgpt}.
In principle, a spoken query that preserves semantic meaning should yield the same visually grounded decision as its textual counterpart. However, modality is not merely a neutral delivery channel \cite{TPAMI,li2023multimodalfoundationmodelsspecialists}. The pathway through which language enters the system may reshape how linguistic signals interact with visual evidence. This raises a central question: \emph{are grounded decisions consistent when equivalent inputs are presented across different modalities and languages?}

Recent work shows that speech-capable systems can exhibit hallucination-like failures, including fluent ASR transcripts weakly grounded in audio \cite{koenecke2024careless,frieske2024asrhallucinations}, increased hallucination under distribution shift \cite{atwany-etal-2025-lost}, and cross-modal grounding failures in audio-visual LLMs that ignore audio and rely on visual priors \cite{nishimura2024audio,sungavhbench}. However, most studies evaluate hallucination \emph{within a single channel} (speech-to-text) or focus on audio-visual description tasks. Our work instead asks whether \emph{visually grounded decisions remain consistent} when the \emph{same intended statement} is delivered via text \textit{vs.} speech across languages.

Evaluating behavioral consistency across modality requires more than aggregate accuracy \cite{srivastava2022beyond}. A model may answer many queries correctly while still exhibiting instability in how it distinguishes supported from unsupported interpretations~\cite{ribeiro-etal-2020-beyond}. To probe consistency directly, we adopt \textbf{a contrastive evaluation design} in which each image is paired with one visually supported statement and two plausible but unsupported alternatives. This structure requires the model to make coherent distinctions within a semantically matched set, rather than succeed on isolated statements. By evaluating performance at the level of the triplet, we can assess whether equivalent inputs lead to consistent grounded decisions across modalities and languages.

\begin{figure}[t]
    \centering
    \includegraphics[width=\linewidth]{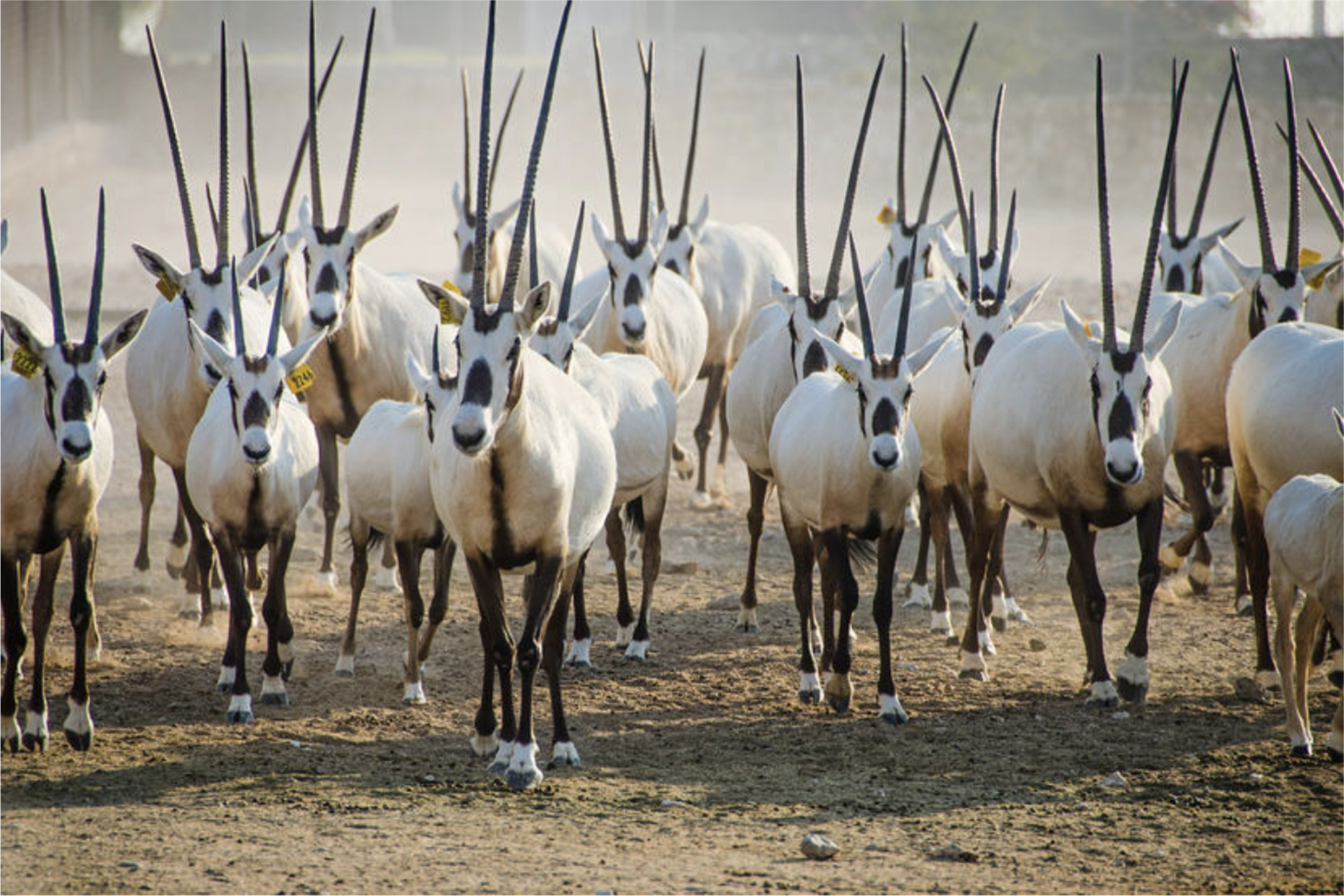}
    \vspace{-4mm}
    \caption{
\textbf{Q$^+$ (\faVolumeUp)} The image shows national animal of Qatar. \\
\textbf{Q$^-$ (\faVolumeUp)} The image shows national animal of Morocco. \\
\textbf{Q$^-$ (\faVolumeUp)} The image shows national animal of Algeria.
}
    \label{fig:oryx_example}
    \vspace{-4mm}
\end{figure}

To our knowledge, there is no publicly available dataset that supports \textit{contrastive evaluation} in an image-grounded multimodal setting spanning text and speech in Arabic and English. We therefore construct a culturally grounded multimodal dataset spanning 18 countries across the Middle East and North Africa (MENA). The dataset covers visually distinctive themes including architecture, traditional clothing, cuisine, religious settings, and public spaces. This diversity enables the creation of semantically plausible yet visually unsupported alternatives, ensuring that distractors remain contextually coherent rather than trivial negatives. For each image, we generate one visually supported statement and two plausible alternatives, forming the contrastive triplets used in our analysis. An example is shown in Figure~\ref{fig:oryx_example}.

\textbf{Contrastive Instability} To quantify decision consistency, we evaluate performance at the triplet level. A model achieves full consistency only if it answers all three statements correctly. However, models may exhibit partial success, resolving some statements correctly while failing others. We therefore define contrastive instability as the rate at which a model fails to achieve full consistency among triplets where it answers at least one statement correctly. This conditional formulation isolates fragmented reasoning from complete failure and reveals instability that may remain hidden under strong aggregate accuracy.


We evaluate recent multimodal foundation models namely: Qwen2.5 (3B and 7B Omni), Qwen3-30B-Omni~\cite{xu2025qwen3omnitechnicalreport}, and Phi-4-multimodal-instruct~\cite{Abdin2024Phi4}, all supporting both text and speech inputs. For speech evaluation, we synthesize English and Arabic queries, enabling controlled comparisons across modality, language, and speaker variation. We also test robustness under speech perturbations to assess stability under realistic acoustic conditions. Our findings reveal:
\begin{itemize}
    \item Speech inputs increase triplet-level instability that is not fully captured by aggregate Q$^+$ and Q$^-$ accuracy, indicating degraded contrastive discrimination rather than performance.
    \item Instability is substantially higher in Arabic than in English, and speech amplifies this cross-lingual gap.
    \item Model scaling improves stability and robustness to acoustic noise, but does not eliminate cross-modal or cross-lingual inconsistency; joint speech--text input mitigates instability without fully restoring text-only coherence.
\end{itemize}

\noindent
We make the following contributions:
\begin{itemize}
    \item We introduce a speech-augmented culturally grounded contrastive benchmark of \textit{10{,}150 triplets}, enabling evaluation of consistency across modality and language.
    \item We propose contrastive instability (CI), a triplet-level metric isolating decision incoherence beyond aggregate accuracy.
    \item We provide a cross-modal, cross-lingual evaluation of multimodal foundation models under text and speech, showing modality is not a neutral input channel.
\end{itemize}

\section{Dataset}
\label{sec:data}



We curated a culturally grounded multimodal image dataset spanning 18 Arab countries, building on the OASIS image collection and M2CQA, to support cross-modal consistency evaluation \cite{alam2025everydaymmqa,mousi2026}. The dataset covers visually distinctive themes including architecture, traditional clothing, cuisine, religious settings, and public spaces, reflecting the importance of cultural context in multimodal understanding and evaluation \cite{nayak-etal-2024-benchmarking,vayani2025all}. Images were collected from the open web using country-specific queries derived from a predefined taxonomy in a human-in-the-loop setup. Retrieval employed geolocalized search settings, image quality constraints, and Creative Commons license filtering, followed by exact and embedding-based near-duplicate removal to ensure diversity and regional appropriateness.

\subsection{Contrastive Triplet Construction} 

For each image in the curated repository, we generate a multiple-choice question (MCQ) consisting of one visually supported answer and two culturally plausible distractors within the same thematic category. 
To ensure that items genuinely require visual grounding, we apply an image-blind filtering step using strong language models and discard MCQs that can be answered without the image, as they are likely solvable from linguistic or cultural priors alone. Each retained MCQ is converted into a contrastive triplet by rewriting its options as standalone statements: the correct answer becomes a visually supported statement (Q$^+$), and the distractors become visually unsupported alternatives (Q$^-$). The statements are minimally contrastive, differing along fine-grained visual or cultural attributes to ensure reliance on image evidence rather than lexical cues (see example in Figure~\ref{fig:oryx_example}).

\subsection{Human Verification and Annotation Reliability}

To assess the reliability of the contrastive annotations, we conduct a human verification study on a randomly sampled subset comprising 15\% of the dataset. The subset is stratified to ensure proportional coverage across countries and taxonomic categories, with images and their corresponding triplet statements sampled uniformly from each country. Annotators are presented with an image and its associated statements and assign two independent labels: \textit{(i)} \textit{Need-Image}, indicating whether answering the statement requires access to the image (\textit{Yes}, \textit{No}, \textit{Unsure}), and \textit{(ii)} \textit{Q-Correctness}, indicating whether the statement is visually supported by the image (\textit{Correct}, \textit{Incorrect}, \textit{Unsure}).

\begin{table}[t]
\centering
\small
\begin{tabular}{lcccc}
\toprule
 & \multicolumn{2}{c}{\textbf{Need-Image}} & \multicolumn{2}{c}{\textbf{Q-Correctness}} \\
\cmidrule(lr){2-3} \cmidrule(lr){4-5}
\textbf{Statement} & \textbf{AC1} & \textbf{Raw Agr.} & \textbf{AC1} & \textbf{Raw Agr.} \\
\midrule
Q$^+$  & 0.996 & 0.996 & 0.752 & 0.784 \\
Q$^-$  & 0.997 & 0.997 & 0.775 & 0.803 \\
Q$^-$  & 0.996 & 0.996 & 0.695 & 0.738 \\
\midrule
Avg    & 0.996 & 0.996 & 0.741 & 0.775 \\
\bottomrule
\end{tabular}
\caption{Inter-annotator agreement for Need-Image and Q-Correctness labels. We report AC1 and percent agreement.}
\label{tab:agreement}
\vspace{-6mm}
\end{table}

In cases of disagreement between annotators, a third annotator performs adjudication. We report both raw percent agreement and Gwet’s AC1 coefficient \cite{gwet2008computing} to account for potential class imbalance. Table~\ref{tab:agreement} summarizes the results. Agreement is near-perfect for the Need-Image label (AC1 $\approx$ 0.996; raw agreement $\approx$ 0.996), indicating consistent identification of statements that require visual evidence. For Q-Correctness, agreement is substantial overall (AC1 $\approx$ 0.741; raw agreement $\approx$ 0.775), with slightly higher consistency for Q$^+$ statements than for Q$^-$ alternatives. Bootstrap 95\% confidence intervals computed via item-level resampling are strictly above zero across conditions, confirming reliability well above chance.

Beyond aggregate agreement statistics, we examine disagreement cases to understand annotation variance. Qualitative inspection shows most disagreements involve visually subtle counterfactual distinctions or fine-grained cultural interpretations rather than systematic errors. Instances of image insufficiency are rare, consistent with the high Need-Image agreement. These findings indicate that the dataset provides a reliable foundation for evaluating triplet-level contrastive consistency.
\subsection{Cross-Lingual and Spoken-Modality Construction}

To evaluate grounded consistency across language and modality, we translate all English statements into Arabic using an in-house machine translation system \cite{fanarteam2026fanar20arabicgenerative}. Translation is performed at the statement level to preserve declarative structure and semantic equivalence of the contrastive triplets, avoiding paraphrasing or lexical restructuring that could alter the Q$^+$/Q$^-$ relationship. We then synthesize spoken versions of each statement in English and Arabic using zero-shot neural speech synthesis,\footnote{\url{https://huggingface.co/coqui/XTTS-v2}} conditioned on natural reference recordings from human speakers \cite{ali2026menaspeechbank}. This setup enables prosodically natural speech while avoiding artifacts associated with fixed proprietary TTS personas. For each statement, we generate matched male and female voice variants, keeping the lexical content strictly identical to the original text (no reformulation or punctuation changes that affect meaning), ensuring speech remains a modality-preserving transformation. To evaluate robustness under realistic acoustic conditions, we further augment the synthesized speech with controlled perturbations, including non-stationary background noise and synthetic reverberation to simulate diverse physical environments. This stress test allows us to determine whether modality-induced instability persists beyond idealized clean audio.

\section{Contrastive Instability}
\label{sec:metric}

Evaluating grounded reasoning using independent statements does not test whether a model can distinguish between closely competing interpretations. A model may correctly accept a visually supported statement in isolation while also accepting a plausible but unsupported alternative. Such behavior reflects inconsistent discrimination rather than coherent grounding.

Contrastive triplets explicitly introduce minimal semantic competition by pairing one visually supported statement ($Q^+$) with two plausible but unsupported alternatives ($Q^-_1$, $Q^-_2$). A triplet is considered \emph{fully consistent} if the model correctly accepts $Q^+$ and rejects both $Q^-_1$ and $Q^-_2$. Triplet accuracy measures the proportion of such fully correct triplets. However, triplet accuracy alone does not reveal how errors arise. A decrease in triplet accuracy may reflect complete misunderstanding (all statements incorrect) or partial inconsistency (some correct, but not all). These behaviors correspond to different failure modes. To isolate fragmented reasoning, we define \textbf{contrastive instability (CI)}. Let $N_{\text{partial}}$ denote the number of triplets with at least one correct statement, and let $N_{\text{consistent}}$ denote the number of those triplets that are fully consistent. We define:
\vspace{-1mm}
\begin{equation}
CI = 1 - \frac{N_{\text{consistent}}}{N_{\text{partial}}}
\nonumber
\end{equation}

\noindent CI measures how often a model fails to resolve a triplet coherently given that it demonstrates at least partial success. Importantly, CI is not a monotonic transformation of triplet accuracy. Two models may exhibit identical triplet accuracy yet differ substantially in CI if one fails primarily through global errors while the other fails through internal inconsistency. By conditioning on partial success, CI separates decision-boundary instability from overall performance degradation. In cross-modal evaluation, this distinction is crucial. Modality shifts may preserve aggregate statement accuracy while increasing internal inconsistency across alternatives. Contrastive instability thus provides a complementary and more diagnostic view of grounded decision behavior across speech and text inputs.





\section{Results}

\textbf{Setup}  We evaluate Qwen2.5-Omni-3B, Qwen2.5-Omni-7B, Qwen3-Omni-30B-A3B-Instruct, and Phi-4-multimodal-instruct using the vLLM-Omni framework.\footnote{https://docs.vllm.ai/projects/vllm-omni/en/latest}
 We consider two settings: \textit{(1)} comparing text versus audio inputs while keeping task instructions textual, and \textit{(2)} evaluating modality effects under joint signaling. All models are prompted to decide whether a statement is True or False using a constrained response format (“The final answer is: <True/False>”). All experiments use greedy decoding for reproducibility, and code and data are released.\footnote{\url{https://github.com/baselmousi/cfhr-ci}}

\begin{table}[t]
\centering
\small
\begin{tabular}{l l l c c c c}
\toprule
\textbf{Model} & \textbf{Lg} & \textbf{Mode} 
& \textbf{Q$^+$} $\uparrow$ 
& \textbf{Q$^-$} $\uparrow$ 
& \textbf{F1} $\uparrow$ 
& \textbf{CI} $\downarrow$ \\
\midrule

Q2.5-3B 
& en & Text & 0.93 & 0.91 & 0.92 & 0.20 \\
& en & Speech & 0.93 & 0.90 & 0.91 & 0.22 \\
& ar & Text & 0.66 & 0.92 & 0.77 & 0.43 \\
& ar & Speech & 0.35 & 0.94 & 0.51 & 0.71 \\
\midrule

Q2.5-7B 
& en & Text & 0.94 & 0.92 & 0.93 & 0.17 \\
& en & Speech & 0.95 & 0.92 & 0.93 & 0.18 \\
& ar & Text & 0.90 & 0.82 & 0.86 & 0.35 \\
& ar & Speech & 0.88 & 0.62 & 0.73 & 0.63 \\
\midrule

Q3-30B 
& en & Text & 0.93 & 0.91 & 0.92 & 0.20 \\
& en & Speech & 0.92 & 0.92 & 0.92 & 0.19 \\
& ar & Text & 0.92 & 0.86 & 0.89 & 0.28 \\
& ar & Speech & 0.92 & 0.73 & 0.81 & 0.47 \\
\midrule

Phi-4M 
& en & Text & 0.57 & 0.94 & 0.71 & 0.50 \\
& en & Speech & 0.61 & 0.83 & 0.70 & 0.58 \\
& ar & Text & 0.45 & 0.77 & 0.57 & 0.75 \\
& ar & Speech & 0.04 & 0.96 & 0.08 & 0.97 \\
\bottomrule
\end{tabular}
\caption{Triplet-level performance across language and modality.  Q denotes Qwen models. 
$\uparrow$ indicates higher is better; $\downarrow$ indicates lower is better. 
CI denotes contrastive instability.}
\label{tab:main_results}
\vspace{-6mm}
\end{table}

\textbf{Evaluation} We evaluate grounded decision consistency across modality, language, and model scale using Q$^+$ and Q$^-$ accuracy, F1, and Contrastive Instability (CI). While Q$^+$ and Q$^-$ summarize isolated correctness and F1 captures aggregate performance, CI measures conditional incoherence within contrastive triplets. Table~\ref{tab:main_results} reports results across English and Arabic under text and speech inputs. Based on these findings, we address the following research questions.


\textbf{RQ1: Does CI provide information beyond Q$^+$ and Q$^-$ accuracy?} 
Q$^+$ and Q$^-$ accuracies summarize aggregate correctness, but they do not capture whether a model coherently distinguishes supported interpretations from closely competing alternatives. As shown in Table~\ref{tab:main_results}, there are cases where aggregate accuracy remains relatively strong while CI remains high, revealing instability that is not visible from statement-level metrics alone. For example, Q2.5-3B in Arabic text achieves an F1 score of 0.77, yet its contrastive instability (CI) is 0.43. This indicates that although the model often answers individual statements correctly, it frequently fails to resolve the full triplet consistently. Such conditional incoherence is not visible from Q$^+$/Q$^-$ accuracy alone. More generally, Table~\ref{tab:main_results} reveals settings in which aggregate correctness remains high while internal decision consistency degrades. These observations demonstrate that standard accuracy metrics do not fully characterize grounded stability. CI provides a complementary diagnostic measure that isolates contrastive instability within semantically matched triplets.


\textbf{RQ2: How does decision stability vary between text and speech?} 
Table~\ref{tab:main_results} shows that speech can substantially increase contrastive instability (CI) relative to text, particularly in Arabic and across multiple models. While English results remain largely stable across modalities, Arabic exhibits pronounced instability under speech. For example, Q2.5-3B’s CI rises from $0.43$ in Arabic text to $0.71$ in Arabic speech, indicating that many triplets that are partially correct become internally inconsistent when the same statements are delivered through speech. Similar trends appear across model scales: Q2.5-7B increases from $0.35$ to $0.63$, and Q3-30B from $0.28$ to $0.47$. These changes occur alongside moderate declines in F1, suggesting that speech introduces instability beyond simple accuracy degradation. Overall, these results indicate that speech is not a neutral input channel, and that modality effects are particularly pronounced in cross-lingual settings.

\begin{figure}[t]
    \centering
    \includegraphics[width=\linewidth]{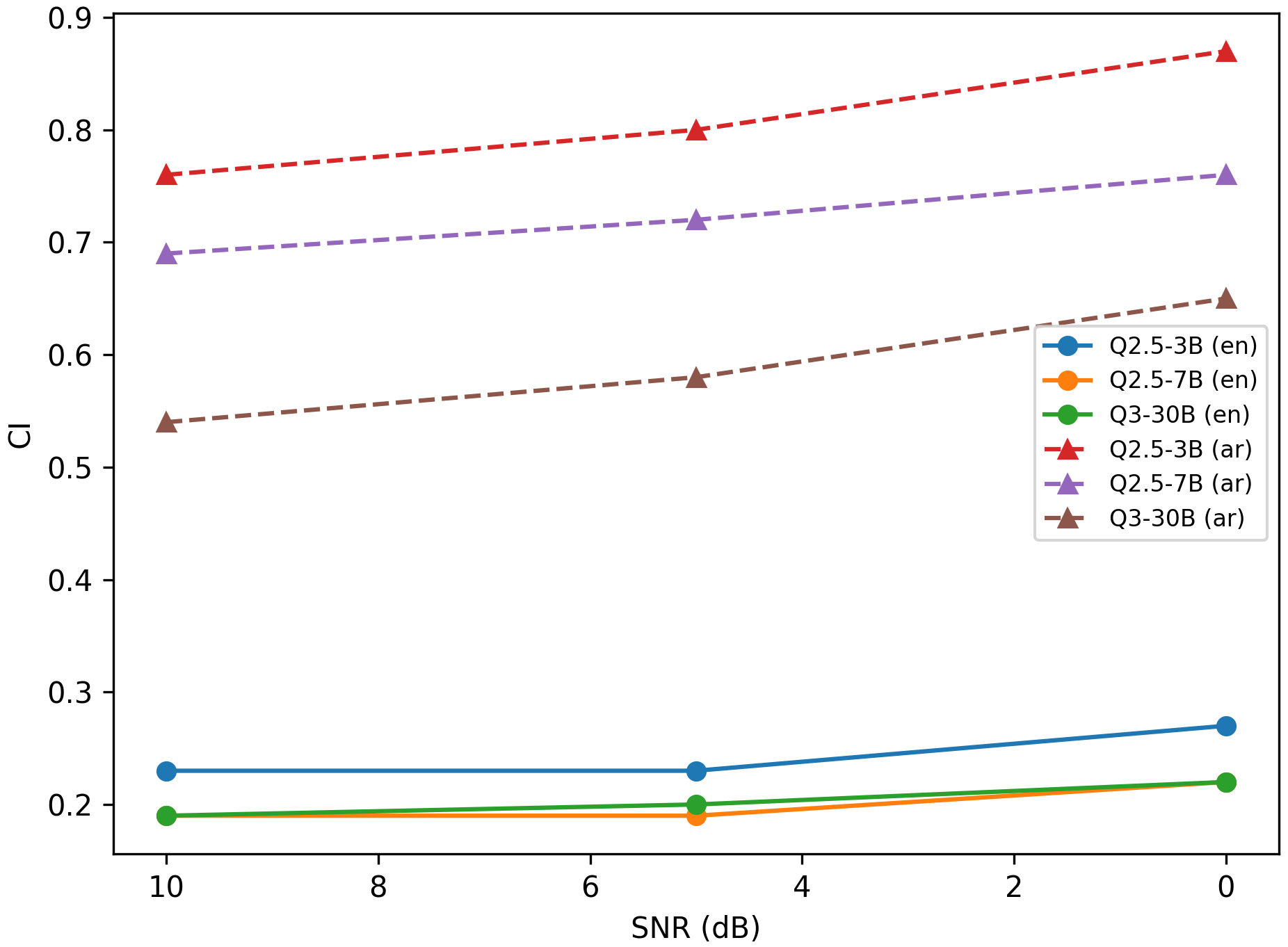}
    \vspace{-4mm}
    \caption{Contrastive instability (CI) as a function of SNR for Qwen models under speech input. Circles (solid) denote English and triangles (dashed) denote Arabic. Lower SNR indicates higher acoustic noise. Results are averaged across voices.}
    \label{fig:robustness}
    \vspace{-4mm}
\end{figure}

 \textbf{RQ3: How does instability vary across languages?} 
Across all models, contrastive instability is consistently higher in Arabic than in English (Table~\ref{tab:main_results}), and the gap widens under speech. For example, for Q2.5-3B under text, CI is $0.20$ in English versus $0.43$ in Arabic; under speech, CI is $0.22$ in English versus $0.71$ in Arabic. Similar gaps appear for larger models (e.g., Q3-30B: $0.20$ vs.\ $0.28$ in text; $0.19$ vs.\ $0.47$ in speech), suggesting that cross-lingual grounding introduces structural instability that is further amplified by spoken input.

 \textbf{RQ4: Does model scaling improve grounded stability?} 
Contrastive instability generally decreases with model size, particularly in Arabic (Table~\ref{tab:main_results}). 
Under text inputs, Arabic CI drops from $0.43$ for Q2.5-3B to $0.35$ for Q2.5-7B and further to $0.28$ for Q3-30B, indicating improved discrimination between supported and unsupported statements as model capacity increases. 
A similar trend appears under speech input, where CI decreases from $0.71$ to $0.63$ and $0.47$ across the same models.  In English, instability remains comparatively low and largely stable across scales (e.g., CI $\approx 0.19$--$0.22$).  Overall, scaling improves robustness and reduces contrastive instability, but does not fully eliminate cross-modal or cross-lingual inconsistency.

\begin{figure}[t]
    \centering
    \includegraphics[width=\linewidth]{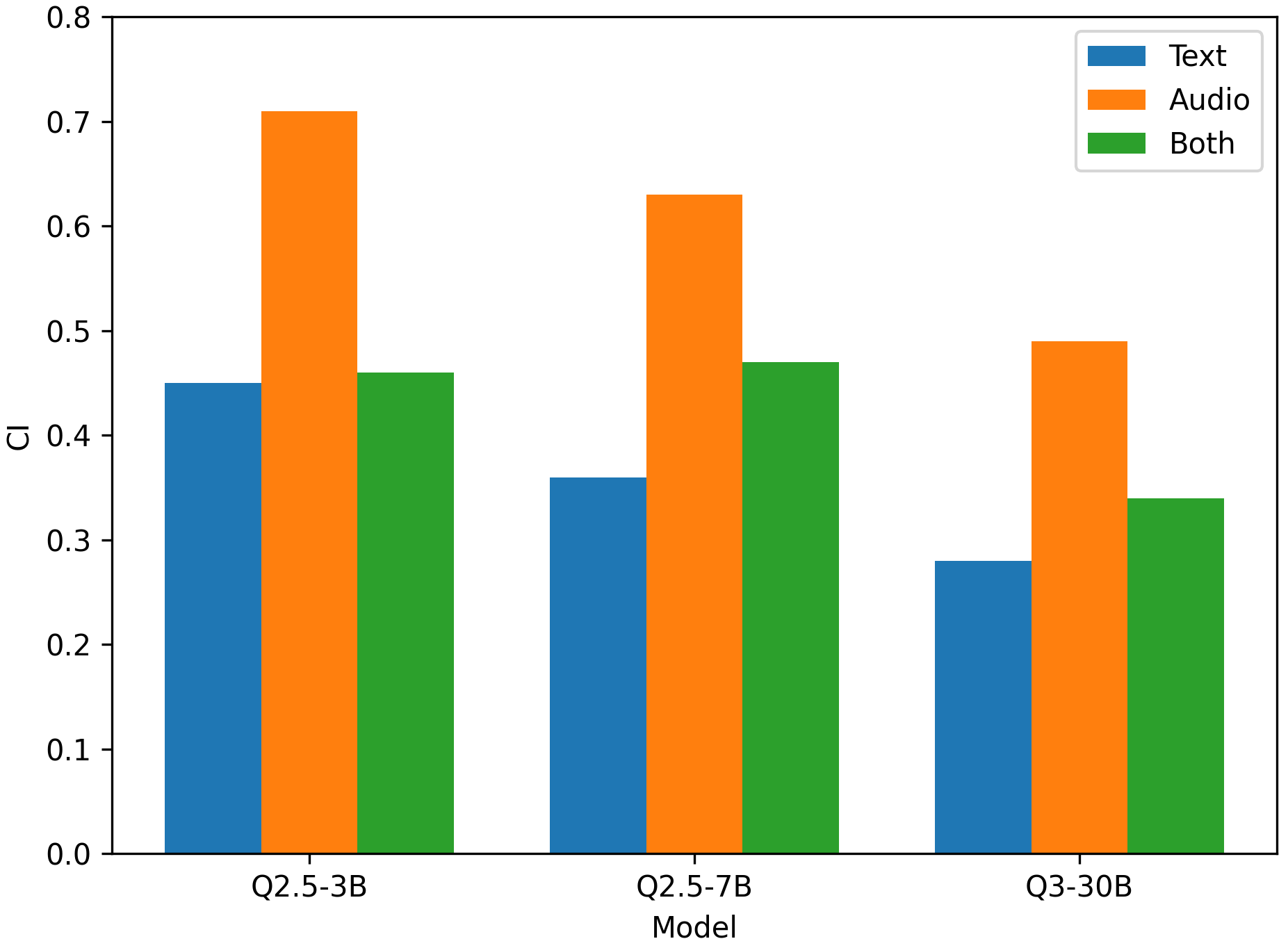}
    \vspace{-4mm}
    \caption{Contrastive instability (CI) under text-only, audio-only, and joint speech–text input for Qwen models in Arabic. Joint signaling reduces instability relative to audio-only input, but does not fully recover text-only stability.}
    \label{fig:joint_signaling}
    \vspace{-5mm}
\end{figure}

\textbf{RQ5: How robust are models to acoustic degradation under speech input?} Figure~\ref{fig:robustness} reports contrastive instability (CI) as a function of SNR for Qwen models under speech input. CI increases monotonically as SNR decreases, indicating that acoustic noise exacerbates grounded decision instability. In English, the degradation is gradual, suggesting robustness under additive noise. In Arabic, the effect is markedly stronger, with sharper CI increases as SNR decreases, indicating that acoustic degradation compounds cross-lingual instability. Model scaling improves robustness, as Q3-30B maintains lower CI across SNR levels. However, the language gap persists even at higher SNRs, suggesting that noise amplifies pre-existing cross-lingual instability rather than causing it.

 \textbf{RQ6: Does joint speech--text signaling reduce instability?} We also evaluate a joint-input setting where models receive both speech and text versions of the same statement. Figure~\ref{fig:joint_signaling} shows contrastive instability (CI) for Qwen models in Arabic under text-only, audio-only, and joint input. Audio-only yields the highest instability, text-only the lowest, and joint input consistently reduces CI relative to audio-only, indicating a stabilizing effect from the textual signal. However, joint input does not fully recover text-only stability, particularly for smaller models. In English (not shown), differences across input conditions are minimal. Overall, multimodal redundancy mitigates but does not eliminate cross-lingual speech-induced instability.


\section{Related Work}


Recent work on multimodal evaluation has shown that aggregate accuracy alone is insufficient to characterize model reliability, motivating behavioral testing and contrastive probing frameworks that assess robustness under controlled variation \cite{ribeiro-etal-2020-beyond,gardner-etal-2020-evaluating,li-etal-2025-treble}. Multimodal benchmarks such as MME \cite{fu2023mme}, SEED-Bench \cite{Li_2024_CVPR} and Hal-Eval \cite{hal_eval} evaluate grounded correctness across diverse tasks, yet primarily report independent-sample performance rather than decision coherence. In multilingual settings, culturally grounded benchmarks and dialect-aware models \cite{nayak-etal-2024-benchmarking, vayani2025all, alwajih-etal-2024-dallah,mousi-etal-2025-aradice} demonstrate that linguistic and regional variation can affect visual grounding behavior. Parallel work on speech-enabled systems highlights robustness challenges under transcription noise and distribution shift \cite{zhang-etal-2023-speechgpt,rubenstein2023audiopalmlargelanguagemodel}, but does not examine whether semantically equivalent inputs delivered via text \textit{vs.} speech yield consistent grounded decisions. Our work addresses this gap by introducing a conditional triplet-level measure of cross-modal instability that isolates internal discrimination failures across modality and language.

\section{Conclusion}

We examined whether multimodal foundation models make consistent visually grounded decisions when semantically equivalent inputs are presented through speech and text. Using a culturally grounded contrastive benchmark and introducing contrastive instability (CI), a conditional triplet-level measure of internal decision incoherence, we show that modality is not a neutral channel of input. Speech often preserves aggregate Q$^+$ and Q$^-$ accuracy while increasing triplet-level instability, particularly in Arabic and under acoustic degradation. Although model scaling and joint speech--text signaling improve robustness, they do not fully eliminate cross-modal or cross-lingual instability. These findings highlight the importance of contrastive evaluation for diagnosing grounded reasoning stability beyond standard accuracy metrics.

\section{Use of Generative AI Disclosure}

Generative AI tools were used in data construction, including question generation, translation, and speech synthesis, and for writing assistance. All research design, analysis, and conclusions are the authors' own..

\bibliographystyle{IEEEtran}
\bibliography{mybib}





















\end{document}